%% file: iclr2026_conference_camera_ready.tex
\documentclass{article} 
\usepackage{iclr2026_conference,times}

\input{math_commands.tex}

\usepackage{xspace}
\usepackage{hyperref}
\usepackage{url}
\usepackage{algorithm}
\usepackage{algorithmic}
\usepackage{booktabs}
\usepackage{multirow}
\newcommand{\eg}{\textit{e.g.}\@\xspace}
\newcommand{\ie}{\textit{i.e.}\@\xspace}
\newcommand{\etc}{\textit{etc.}\@\xspace}

\usepackage{hyperref}
\usepackage{url}
\usepackage{graphicx}

\usepackage{wrapfig}
\usepackage{amsthm}
\newtheorem{proposition}{Proposition}

\usepackage[table]{xcolor}
\title{SoftCFG: Uncertainty-guided Stable Guidance for Visual autoregressive Model}

\author{Dongli Xu \thanks{Corresponding Author} \\
Processing Speech and Images, ESAT\\
KU Leuven\\
Leuven, Belgium \\
\texttt{dongliixu@gmail.com} \\
\And
Aleksei Tiulpin \\
HST Research Unit, Faculty of Medicine\\
University of Oulu \\
Oulu, Finland \\
\texttt{aleksei.tiulpin@oulu.fi} \\
\AND
Matthew B. Blaschko \\
Processing Speech and Images, ESAT \\
KU Leuven\\
Leuven, Belgium \\
\texttt{matthew.blaschko@esat.kuleuven.be}
}

\iclrfinalcopy 

\begin{document}

\maketitle
\begin{abstract}
Autoregressive (AR) models have emerged as powerful tools for image generation by modeling images as sequences of discrete tokens. 
While Classifier-Free Guidance (CFG) has been adopted to improve conditional generation, its application in AR models faces two key issues: \emph{guidance diminishing}, where the conditional–unconditional gap quickly vanishes as decoding progresses, and \emph{over-guidance}, where strong conditions distort visual coherence. 
To address these challenges, we propose \textbf{SoftCFG}, an uncertainty-guided inference method that distributes adaptive perturbations across all tokens in the sequence. 
\textit{The key idea behind SoftCFG} is to let each generated token contribute certainty-weighted guidance, ensuring that the signal persists across steps while resolving conflicts between text guidance and visual context. 
To further stabilize long-sequence generation, we introduce \textbf{Step Normalization}, which bounds cumulative perturbations of SoftCFG. 
Our method is training-free, model-agnostic, and seamlessly integrates with existing AR pipelines. 
Experiments show that SoftCFG significantly improves image quality over standard CFG and achieves state-of-the-art FID on ImageNet $256\times256$ among autoregressive models.

\begin{figure}[h]
\centering
\includegraphics[width=\linewidth]{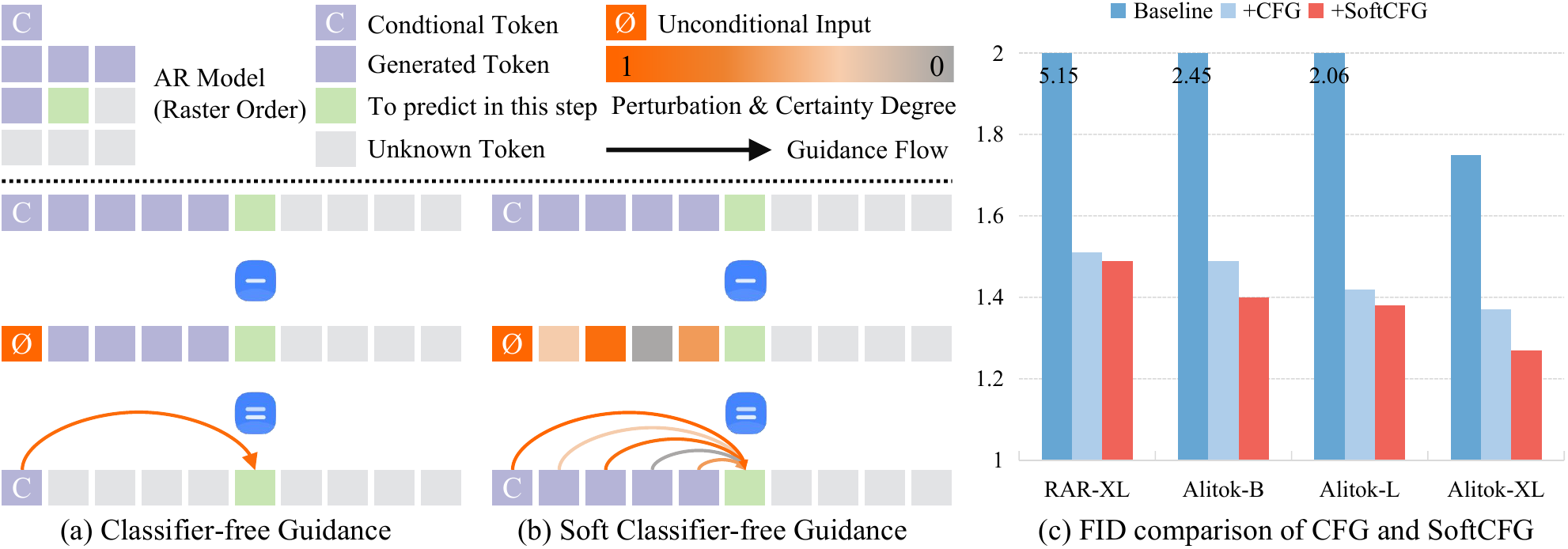}
\vspace{-0.6cm}
    \caption{Comparison between standard CFG and proposed SoftCFG. 
(a) Standard CFG in AR models perturbs only the first class token by replacing it with an empty token. 
(b) SoftCFG adaptively adjusts perturbation strength according to model uncertainty, providing smoother and more informative guidance. 
(c) Quantitative results on ImageNet $256\times256$ show that SoftCFG consistently reduces FID across diverse AR models compared to baseline sampling and standard CFG.
}    \label{fig:comparison_cfg_softcfg}

\end{figure}

\end{abstract}

\begin{figure}
    \centering
\includegraphics[width=1\linewidth]{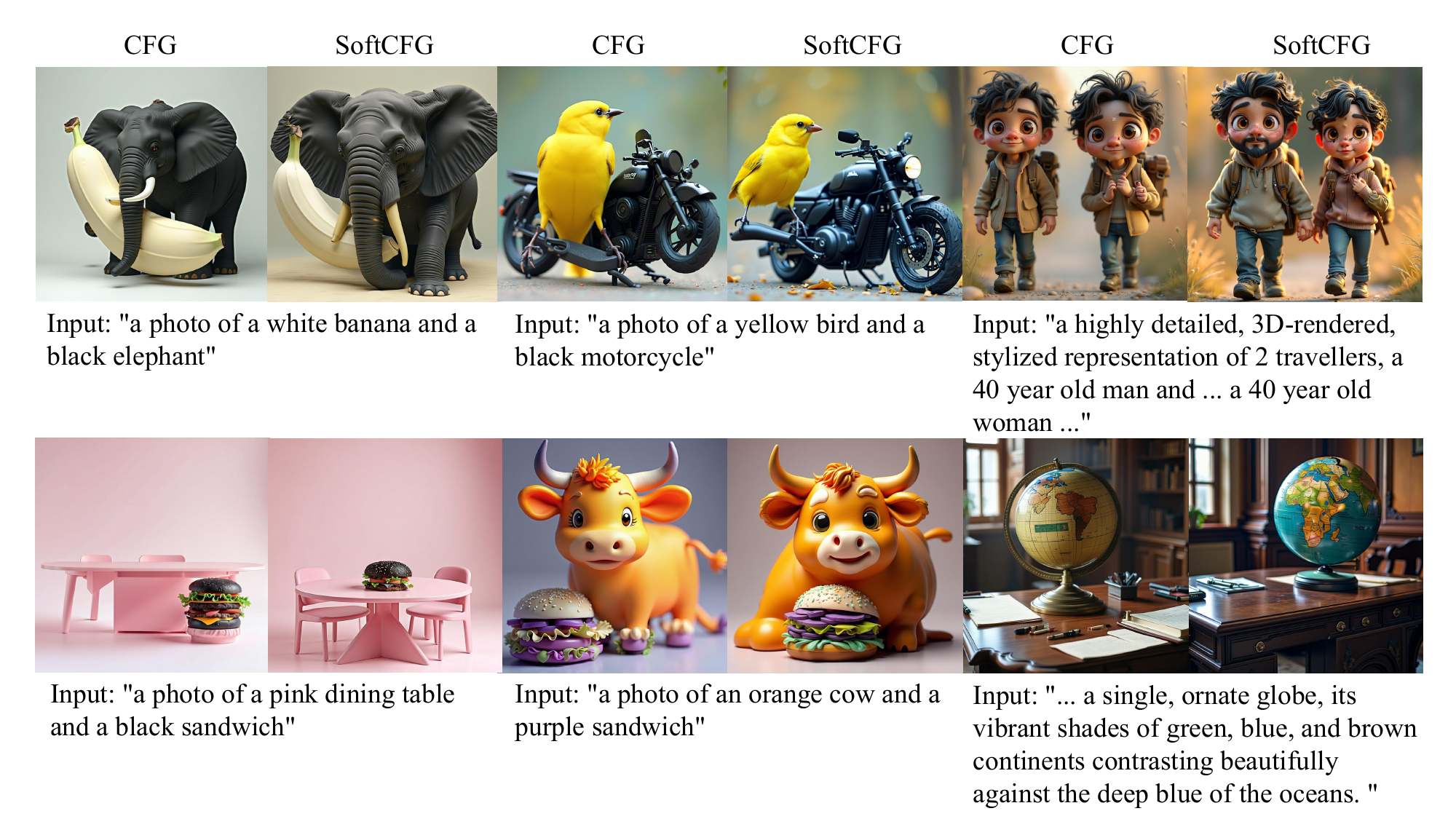}
    \caption{Comparison of images generated by standard Classifier-Free Guidance (CFG) and our proposed SoftCFG on LuminaGPT-8B~\cite{xin2025luminamgpt}. Unlike CFG, which applies the same conditional offset regardless of generation history, SoftCFG adaptively incorporates uncertainty from the already generated content. As a result, SoftCFG effectively reduces unreasonable artifacts, such as motorcycles collapsing into tangled shapes, extra trunks emerging from nowhere, or redundant hands in humans. This demonstrates that by aligning guidance with generated content, SoftCFG yields more coherent and visually plausible generations.}
\vspace{-0.3cm}
\label{fig:comparison_cfg_softscfg}
\end{figure}

\section{Introduction}

\begin{wrapfigure}{t}{0.6\textwidth} 
\vspace{-0.6cm}
\includegraphics[width=\linewidth]{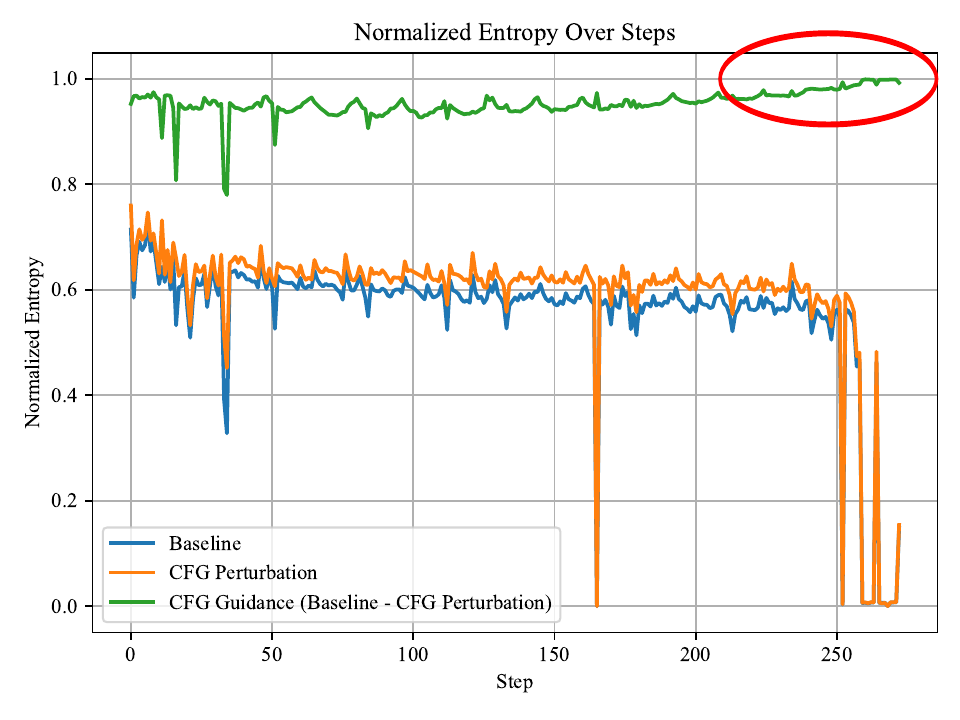} 
  \caption{Diminishing effect of classifier-free guidance (CFG) in AR model Alitok-XL~\cite{wu2025alitok}. We plot the normalized entropy over generation steps. As generation progresses, the difference~(\ie, the guidance signal, green line in the plot) between baseline~(blue line) and CFG perturbation~(orange line) entropy quickly vanishes. Here, a normalized entropy close to 1 indicates that guidance no longer provides informative guidance, please refer to Appendix~\ref{app:entropy} for more details of normalized entropy.}
  \vspace{-1cm}
\label{fig:dimnishing_entropy}
\end{wrapfigure}
Visual Autoregressive~(AR) models~\citep{van2016pixelcnn, chang2022maskgit, sun2024llamgen} formulate image generation as a next-token prediction task over sequences of discrete visual tokens, typically obtained via vector-quantized autoencoders~\citep{van2017neural}.
Given a sequence of previously generated tokens, a visual AR model predicts the next one with a decoder-only transformer, following the same autoregressive formulation that underlies large language models~(LLMs)~\citep{touvron2023llama}.
This unified design enables conceptual and architectural alignment between vision and language, offering simplicity, scalability, and potential for cross-modal integration~\citep{wu2025janus,xie2024show-o,xin2025luminamgpt}.

Despite these advances, the quality of AR-based image generation is still heavily influenced by the inference process.
Recent work~\citep{sun2024llamgen} has introduced classifier-free guidance~(CFG)~\citep{ho2022cfg} into the AR setting, aiming to improve conditional generation by amplifying the difference between conditional and unconditional predictions during inference.
As illustrated in Fig.~\ref{fig:comparison_cfg_softcfg}~(a), in autoregressive models, CFG is typically applied by generating two parallel predictions at each generation step: one conditioned on the target class (by including a class token or embedding at the start of the input sequence), and one unconditional version where the first class token is replaced with an empty token.
This technique, originally developed for diffusion models, has shown promising results~\citep{yu2024rar,ren2024flowar,wu2025alitok} in steering AR generation towards class conditions without requiring auxiliary classifiers.
\begin{figure}
    \centering
\includegraphics[width=0.8\linewidth]{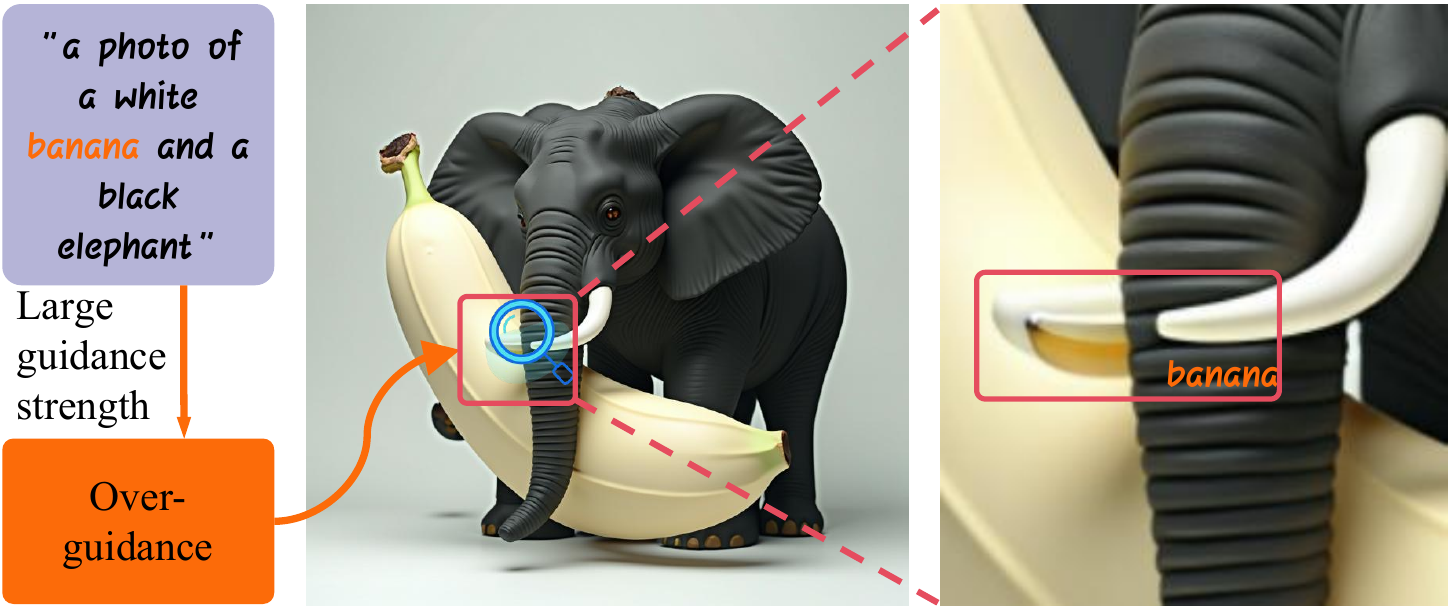}
    \caption{Illustration of the over-guidance phenomenon. 
When applying a large guidance strength, the model over-emphasizes certain words in the prompt (\eg, \textit{``banana''}), leading to distorted generations. 
In this example, the model incorrectly maps the word ``banana'' to the elephant’s tusk, highlighting how excessive guidance strength can harm semantic alignment.}
    \label{fig:over_guidance}
\end{figure}

However, applying CFG to AR models introduces two fundamental challenges: 
\textbf{First, the guidance signal can \emph{diminish} over time}: in contrast to diffusion, where guidance is injected at every denoising step, AR models rely heavily on conditioning tokens at the beginning of the sequence. As decoding proceeds, these tokens drift away from the local context, causing the conditional–unconditional gap to shrink and eventually vanish, even within short sequences (\eg, $16{\times}16$ grids), as shown in Fig.~\ref{fig:dimnishing_entropy}. Some recent methods~\citep{tian2024var,yu2024rar,han2025infinity} mitigate this by injecting conditional embeddings directly into the prediction head (\eg, AdaLN~\citep{perez2018film}), ensuring conditional information remains accessible. 

\textbf{Second, CFG can also suffer from \emph{over-guidance}.} Since guidance depends solely on external conditions such as class labels or text prompts, increasing the guidance scale often enforces semantics too aggressively, conflicting with visual coherence. This leads to artifacts such as duplicated limbs, redundant handlebars, or spurious object parts, as illustrated in Fig.~\ref{fig:over_guidance}. Even methods that inject conditional embeddings at every step remain tied to external signals and thus cannot fully resolve this semantic–visual conflict. 

Together, this duality mirrors gradient vanishing and explosion in neural network training, which are usually mitigated by normalization and regularization methods. 
Motivated by this analogy, we propose \textbf{SoftCFG}: an uncertainty-guided inference method that distributes guidance more stably across the generation sequence. 
\textit{The key idea behind SoftCFG is} \textbf{to let every generated token contribute certainty-weighted guidance, ensuring the signal persists across steps while naturally regularizing conflicts between text guidance and context}, as illustrated in Fig.~\ref{fig:comparison_cfg_softcfg}~(b).

\begin{figure}[t]
    \centering
\includegraphics[width=1\linewidth]{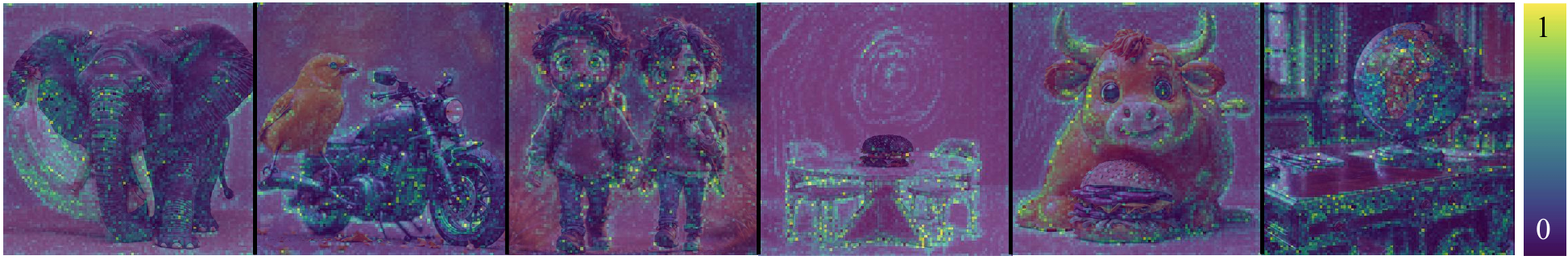}
    \caption{Heatmaps of token confidence overlaid on generated images by LuminamGPT2~\citep{xin2025luminamgpt}. High-confidence regions align well with salient semantic structures (\eg, object parts), while low-confidence regions occur in ambiguous backgrounds, supporting high-confidence tokens can be effective guidance signals.}
    \label{fig:heatmap_confidence}
\end{figure}

In this work, we instantiate the weighting by prediction confidence which can be aligned well with the semantics of generated content~(as illustrated in Fig.~\ref{fig:heatmap_confidence}), but the framework is general and can accommodate learned scorers or perceptual alignment measures. 
Similar to how CFG perturbs the class token to encourage alignment with the class condition, SoftCFG perturbs high-confidence tokens to encourage future tokens to align with the most semantically reliable content generated so far.
Therefore, SoftCFG not only alleviates the fading guidance problem but also reconciles the conflict between semantic alignment and visual coherence, leading to more stable and plausible generations.

\begin{figure}[t]
\vspace{-0.3cm}
\centering
\includegraphics[width=\linewidth]{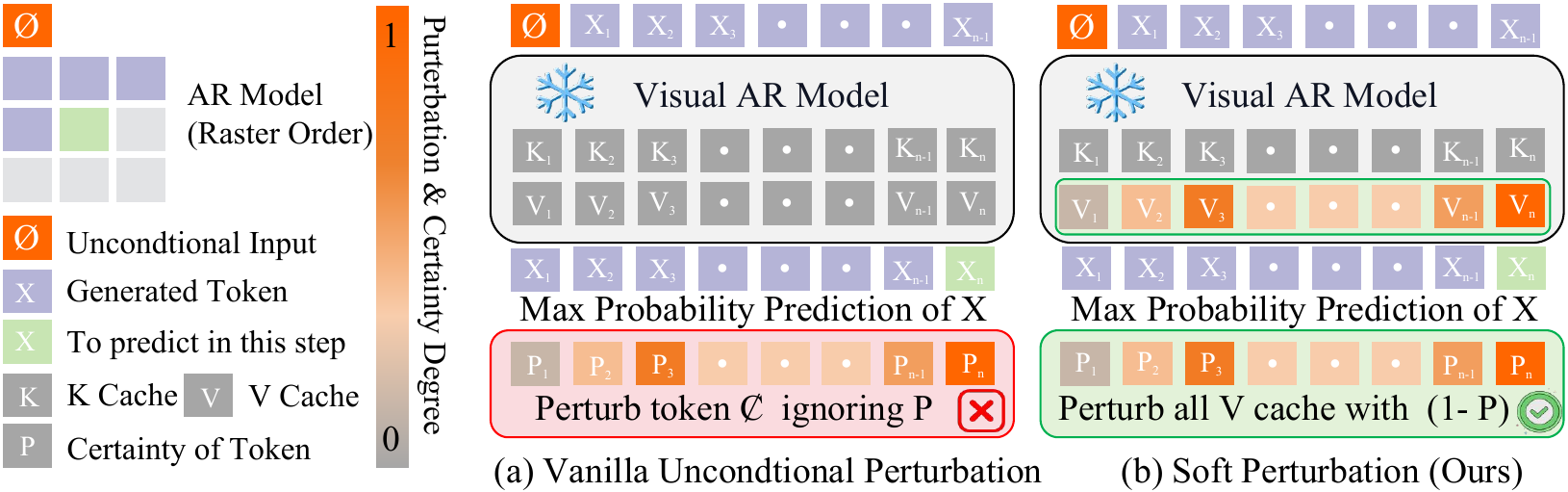}
\vspace{-0.5cm}
    \caption{Two perturbation strategies for Visual AR models.
(a)~Unconditional Perturbation modifies the first class conditional token regardless of the certainty score.
(b)~Uncertainty-guided Perturbation applies softer, weighted changes to all $\mathbf{V}$ cache entries, with strength $(1-\mathbf{P})$, offering stronger perturbation to high-confidence tokens.}
    \label{fig:method}
\vspace{-0.3cm}
\end{figure}

To realize this idea, we need a mechanism that allows generated tokens to directly influence future decoding steps. Since the value cache stores the representations of past tokens that are repeatedly attended to by subsequent predictions, it provides a natural handle for injecting token-wise guidance. We therefore compute each token’s maximum predicted probability $p_{\max}$ as a measure of its confidence, and scale its value cache across all attention layers by a factor of $(1 - p_{\max})$ during inference, as illustrated in \figurename~\ref{fig:method}(b). In this way, tokens with higher confidence receive stronger perturbations on their cached representations, which amplifies their downstream impact and distributes guidance signals throughout the sequence. To avoid degenerate behavior when many tokens accumulate large perturbations or when $p_{\max}$ approaches 1, we further introduce a \textbf{Step Normalization} procedure: at each step, we normalize the set of perturbation weights so that their sum remains constant (\eg, 1), ensuring stable and balanced guidance throughout the generation process.

SoftCFG requires \textbf{no additional training}, introduces \textbf{no architectural change}, and adds \textbf{negligible computational overhead}, making it fully compatible with existing AR models and CFG setups.
By leveraging uncertainty-weighted perturbations and step normalization, it improves both the stability and effectiveness of guidance during inference. 
Experiments on ImageNet show that SoftCFG significantly improves generation quality, reducing the FID from $\mathbf{1.37}$ to $\mathbf{1.27}$ over a SOTA AR baseline~\citep{wu2025alitok}, setting a new state of the art for autoregressive models on $256\times256$~\cite{deng2009imagenet} benchmark.

\noindent We conclude our contributions as follows:
\begin{itemize}
\item We demonstrate the necessity of leveraging generated visual content as guidance for subsequent token prediction, extending the notion of guidance beyond class or text tokens.
\item We propose SoftCFG, a general framework that introduces token-wise soft weighting to integrate such guidance. In this work, we instantiate the weighting by token confidence, while other scoring functions or learned modules can be naturally incorporated.
\item Through comprehensive experiments on class-conditional benchmarks (\eg, ImageNet $256\times 256$), we show that SoftCFG achieves state-of-the-art FID among autoregressive models, consistently improving quality and alignment with negligible overhead.
\end{itemize}

\section{Soft Classifier-free Guidance}

\subsection{Preliminary: Classifier-Free Guidance in AR Models}
\label{sec:cfg}

Classifier-Free Guidance (CFG) is a widely used technique to enhance controllability in generative models by interpolating predictions from conditional and unconditional branches. At each decoding step $t$, an autoregressive model maintains key/value caches $(\mathbf{K}_{<t},\mathbf{V}_{<t})$ summarizing previously generated tokens $x_{<t}$. The branch logits are
\begin{equation}
\mathbf{z}_{t}^{\text{cond}}   = f_\theta(\mathbf{K}_{<t}^{\text{cond}},   \mathbf{V}_{<t}^{\text{cond}},   x_{t-1}, c), \quad
\mathbf{z}_{t}^{\text{uncond}} = f_\theta(\mathbf{K}_{<t}^{\text{uncond}}, \mathbf{V}_{<t}^{\text{uncond}}, x_{t-1}, \emptyset),
\end{equation}
and the guided logits are formed as
\begin{equation}
\label{eq:cfg}
\mathbf{z}^{\text{CFG}}_t = \mathbf{z}^{\text{uncond}}_t + \gamma\,(\mathbf{z}^{\text{cond}}_t-\mathbf{z}^{\text{uncond}}_t), \quad
p_\theta^{\text{CFG}}(\cdot \mid x_{<t}, c) = \mathrm{softmax}(\mathbf{z}^{\text{CFG}}_t),
\end{equation}
where $\gamma>0$ is the guidance scale. Here, we use $\Delta_t=\mathbf{z}^{\text{cond}}_t-\mathbf{z}^{\text{uncond}}_t$ to denote the step-wise guidance offset, which is only decided by input text condition. 

Standard CFG applies a \emph{hard offset} $\gamma\Delta_t$ at every step. While this improves conditional alignment, it can also \emph{over-amplify} the conditional signal, especially when $\|\Delta_t\|$ or $\gamma$ is large. This phenomenon, which we term \emph{over-guidance} as illustrated in Fig.~\ref{fig:over_guidance}, often manifests as structural artifacts (\eg, duplicated parts or implausible object completions).
Moreover, in AR models, once $x_t$ is sampled, this offset does not propagate explicitly to future steps, so the effect of $\Delta_t$ may \emph{decay too quickly} across long sequences, as shown in Fig.~\ref{fig:dimnishing_entropy}. These two limitations motivate the development of a softer, regularized alternative to CFG.

\subsection{SoftCFG: Uncertainty-Guided Perturbation}
\label{sec:softcfg}

At a high level, our goal is to address the over-guidance and guidance diminishing issues identified in Sec.~\ref{sec:cfg}, and make guidance both \emph{context-aware} and \emph{stable} over long horizons.
Instead of relying solely on external conditions, we incorporate already generated content and regularize guidance through the model’s context.
Concretely, SoftCFG acts as a context-aware regularizer on the \emph{unconditional} branch: it softly adjusts internal states so that reliable tokens contribute more and uncertain ones less.
This redistribution embeds guidance into the model’s memory, enabling it to persist across future decoding steps—without retraining or architectural changes.

Specifically, for each previously generated token $i<t$, we compute its predictive confidence using the maximum probability from the conditional distribution:
\begin{equation}
w_i = 1 - p_{\max}(x_i), \qquad p_{\max}(x_i) = \max\nolimits_{v} \; p_\theta^{\text{cond}}(\cdot \mid x_{<i},c).
\end{equation}
Tokens with higher confidence (low $w_i$) are considered more reliable and thus receive stronger perturbations. We also test different confidence indicators in Sec.~\ref{sec:ablations}. We scale their unconditional value vectors as:
\begin{equation}
\tilde{\mathbf{v}}^{\text{uncond, pertcontext}}_i = w_i\,\mathbf{v}^{\text{uncond}}_i,
\end{equation}
where $\tilde{\mathbf{v}}^{\text{uncond, pertcontext}}_i$ denotes this value cache has no conditional information and its context information is also perturbed.
Then we can produce context-perturbed unconditional logits as follows:
\begin{equation}
\tilde{\mathbf{z}}^{\text{uncond, pertcontext}}_t = f_\theta(\mathbf{K}_{<t}^{\text{uncond}}, \tilde{\mathbf{V}}_{<t}^{\text{uncond, pertcontext}}, x_{t-1}, \emptyset).
\end{equation}

SoftCFG then combines the conditional branch with the perturbed unconditional branch:
\begin{equation}
\label{eq:softcfg}
\mathbf{z}^{\text{SoftCFG}}_t 
=\mathbf{z}^{\text{cond}}_t + \gamma\,(\tilde{\mathbf{z}}^{\text{uncond, pertcontext}}_t -\mathbf{z}^{\text{cond}}_t).
\end{equation}
In this way, SoftCFG weakens the unconditional context based on token confidence: reliable tokens are downscaled more, while uncertain ones are preserved.
This adaptive adjustment redefines the guidance itself: instead of relying solely on the external condition, SoftCFG incorporates normalized context from already generated tokens, allowing the guidance signal to be maintained and propagated consistently across future steps.

Equivalently, SoftCFG can be written as a regularized variant of CFG in Eq.~\ref{eq:cfg}:
\begin{equation}
\label{eq:context_aware_regularizer}
\mathbf{z}^{\text{SoftCFG}}_t 
= \mathbf{z}^{\text{CFG}}_t + \gamma\,\Delta^{\text{context}}_t, 
\qquad \Delta^{\text{context}}_t = \tilde{\mathbf{z}}^{\text{uncond, pertcontext}}_t - \mathbf{z}^{\text{uncond}}_t.
\end{equation}
The correction $\Delta^{\text{context}}_t$ is induced by selectively weakening the unconditional value cache, and hence is bounded by the perturbation weights $\{w_i\}$.

Geometrically, the additional term $\Delta^{\text{context}}_t$ acts as a \emph{context-aware regularizer}. 
When $\Delta^{\text{context}}_t$ is aligned with the original guidance offset $\Delta_t=\mathbf{z}^{\text{cond}}_t-\mathbf{z}^{\text{uncond}}_t$, SoftCFG amplifies the conditional signal; when misaligned, it shrinks it. 
This mirrors the role of classical $\ell_2$ regularization, which constrains parameter updates by pulling them back toward the origin. 
Here, instead of parameters, we regularize the \emph{contextual representations}, ensuring that the guidance signal remains bounded and smoothly propagated through the sequence.

\begin{algorithm}[t]
\caption{SoftCFG sampling with Step Normalization}
\begin{algorithmic}[1]
\STATE Initialize $x_0 \gets \langle bos \rangle$; empty caches $\{\mathbf{K}^{\text{cond}},\mathbf{V}^{\text{cond}},\mathbf{K}^{\text{uncond}},\mathbf{V}^{\text{uncond}}, \mathbf{P}_\text{max}\}$
\FOR{$t=1$ to $T$}
  
  \STATE $\mathbf{z}^{\text{cond}}_t \!\gets\! f_\theta(\mathbf{K}^{\text{cond}}_{<t},\mathbf{V}^{\text{cond}}_{<t},x_{t-1},c)$
  \STATE \textbf{for} $i=1,\dots,t\!-\!1$: $w_i \gets 1 - p_{\max}(x_i)$ \hfill $\triangleright$ get the confidence of all generated token
  \STATE \textbf{for} $i=1,\dots,t\!-\!1$: 
$\hat w_i \gets 1-\dfrac{1-w_i}{\sum_{j=1}^{t-1}(1-w_j)+\varepsilon}$ 
\hfill $\triangleright$ step normalization
  \STATE$\tilde{\mathbf{V}}^{\text{uncond}}_{<t} \!\gets\! \mathbf{V}^{\text{uncond}}_{<t}$;\quad \textbf{for} $i=1,\dots,t\!-\!1$: $\tilde{\mathbf{v}}^{\text{uncond}}_i \!\gets\! \hat w_i \cdot \mathbf{v}^{\text{uncond}}_i$ \hfill $\triangleright$ apply soft perturbation
  \STATE $\tilde{\mathbf{z}}^{\text{uncond,pertcontext}}_t \!\gets\! f_\theta(\mathbf{K}^{\text{uncond}}_{<t},\tilde{\mathbf{V}}^{\text{uncond}}_{<t},x_{t-1},\emptyset)$
  \STATE $\mathbf{z}^{\text{SoftCFG}}_t \!\gets\! (1+\gamma)\mathbf{z}^{\text{cond}}_t - \gamma\,\tilde{\mathbf{z}}^{\text{uncond,pertcontext}}_t$ \hfill $\triangleright$ SoftCFG with context-regularization
  \STATE $x_t \sim \mathrm{Sample}\!\big(\mathrm{softmax}(\mathbf{z}^{\text{SoftCFG}}_t)\big)$
  \STATE Update \emph{original} caches with $x_t$ to obtain $\{\mathbf{K}^{\text{cond}}_{\le t},\mathbf{V}^{\text{cond}}_{\le t},\mathbf{K}^{\text{uncond}}_{\le t},\mathbf{V}^{\text{uncond}}_{\le t},\mathbf{P}_\text{max}\}$;
\ENDFOR
\end{algorithmic}
\label{alg:softcfg}
\end{algorithm}

\subsection{Step Normalization}
\label{sec:stepnorm}
\begin{figure}[t]
\vspace{-0.5cm}
    \centering
    \begin{minipage}[t]{0.48\linewidth}
        \centering
    \includegraphics[width=\linewidth]{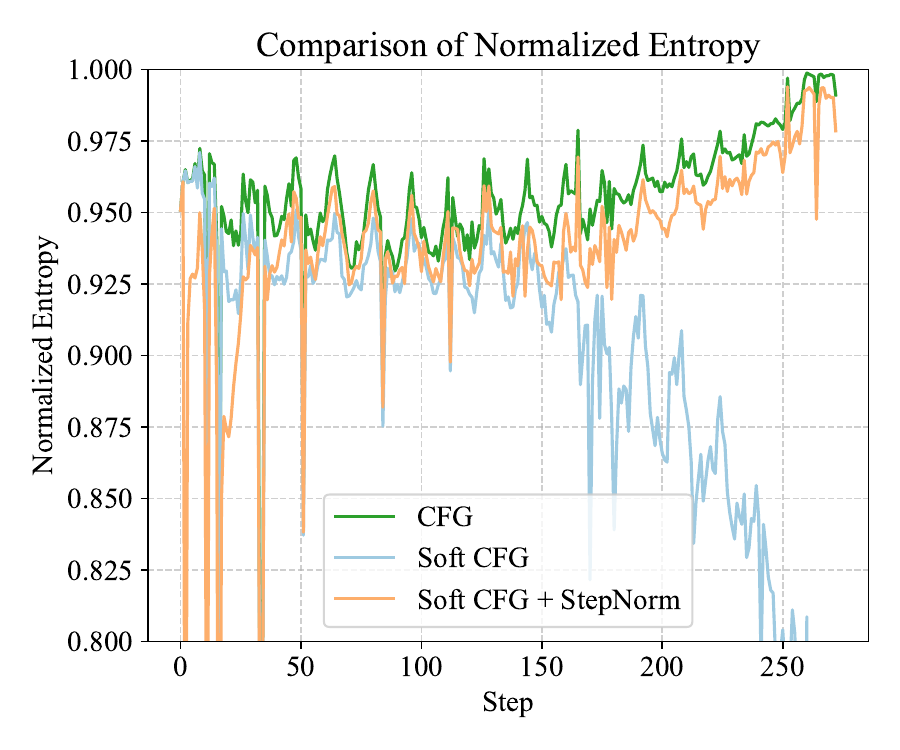}
    \vspace{-0.5cm}
        \caption{Comparison of normalized entropy across different sampling strategies. 
Step normalization stabilizes SoftCFG by controlling cumulative perturbations of the unconditional branch.}
        \label{fig:step_norm_entropy}
    \end{minipage}
    \hfill
    \begin{minipage}[t]{0.48\linewidth}
        \centering
        \includegraphics[width=\linewidth]{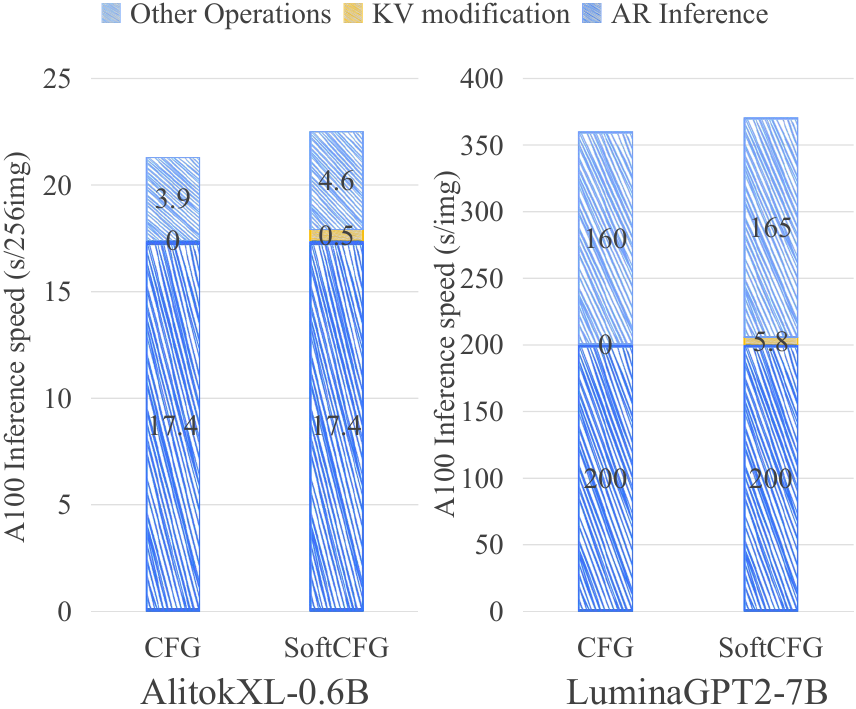}
        \vspace{-0.5cm}
        \caption{Inference speed comparison of CFG and SoftCFG~(w\ Step Norm) on AliTok-0.6B ($256\times256$) and LuminaGPT2-7B ($756\times756$). SoftCFG adds negligible overhead.}
        \label{fig:time_complexity}
    \end{minipage}
\end{figure}

However, we observed an explosion of \(\Delta^{\text{context}}_t\) in Fig.~\ref{fig:step_norm_entropy}, where the normalized entropy of the guidance increases rapidly, indicating that unbounded SoftCFG may cause the guidance to explode. We attribute this to the cumulative growth of perturbations in the unconditional branch.

Formally, the deviation of SoftCFG from vanilla CFG, defined by the context-aware regularizer $\Delta^{\text{context}}_t$ in Eq.~\ref{eq:context_aware_regularizer}, is bounded by:
\begin{equation}
\|\Delta^{\text{context}}_t\| \leq L_t \cdot \sum_{i<t}(1-w_i)\|\mathbf{v}_i^{\text{uncond}}\|,
\end{equation}
where $L_t$ is the Lipschitz constant of $f_\theta$ with respect to the value cache. Please refer to Appendix~\ref{app:cfg_bound} for more details. If \(\sum_{i=1}^{t-1} (1 - w_i)\) were bounded, the deviation would be at most \( O(\gamma) \) within a controlled trust region.

However, as the sequence length \( t \) increases, the cumulative deviation \(\sum_{i=1}^{t-1} (1 - w_i) \) can grow significantly, especially when many tokens have high confidence (\( w_i \to 0 \), \( 1 - w_i \to 1 \)), potentially causing \(\Delta^{\text{context}}_t\) to become excessively large and leading SoftCFG to deviate far from vanilla CFG.
This also explains the degeneration observed in practice: \textit{the unconditional branch gradually loses all contextual information, causing instability in long-horizon generation.}
Formally, if $w_i \to 0$ for high-confidence tokens, the contribution of $\mathbf{V}_{<t}^{\text{uncond}}$ is suppressed and the unconditional branch degenerates:
$\mathbb{E}[\tilde{\mathbf{z}}^{\text{uncond,pertcontext}}_t] \to 0.$

To mitigate this, we introduce \textbf{Step Normalization}, which renormalizes the perturbation weights at every step:
\begin{equation}
\hat{w}_i = 1-\frac{1-w_i}{\sum_{j=1}^{t-1}(1-w_j)} \quad \text{such that} \quad \sum_{i=1}^{t-1}(1-\hat{w}_i) = 1.
\end{equation}

\begin{proposition}[Bounded Deviation of Step-Normalized SoftCFG]
Let $f_\theta$ be $L_t$-Lipschitz with respect to its value-cache input at step $t$.
Then the deviation of SoftCFG from vanilla CFG is bounded as
\begin{equation}
\|\Delta^{\text{context}}_t\| 
= \|\tilde{\mathbf{z}}^{\text{uncond,pert}}_t - \mathbf{z}^{\text{uncond}}_t\|
\leq L_t \cdot \max_{i<t}\|\mathbf{v}_i^{\text{uncond}}\|,
\end{equation}
where $\tilde{\mathbf{z}}^{\text{uncond,pert}}_t$ denotes the unconditional logits under step-normalized perturbation.
\end{proposition}

Without normalization, the cumulative perturbation of past tokens grows proportionally with sequence length.
Step normalization rescales the perturbation weights $\{1-w_i\}$ so that 
$\sum_{i=1}^{t-1}(1-\hat{w}_i)=1$ at each step, effectively allocating a unit perturbation budget over the context.
Therefore the perturbation magnitude at step $t$ is controlled by at most one token’s contribution, leading to the bound
$\|\Delta^{\text{context}}_t\|\leq L_t \cdot \max_{i<t}\|\mathbf{v}_i^{\text{uncond}}\|$.

\section{Experiments}
\subsection{Setup}
\paragraph{Dataset and Benchmark.}
For \emph{class-conditional} generation, we evaluate on ImageNet~\citep{deng2009imagenet} with the standard $256{\times}256$ class-conditional generation protocol.
For \emph{text-to-image} generation, we evaluate on two widely used benchmarks, \ie, GenEval benchmark~\citep{ghosh2023geneval} and DPG-Bench~\citep{hu2024ella}.

\paragraph{Models.}
We evaluate SoftCFG on state-of-the-art (SOTA) autoregressive (AR) generators to ensure that improvements are not due to weak baselines. 
For \emph{class-conditional} generation, we use the strongest published AR model on ImageNet $256{\times}256$~\citep{deng2009imagenet}, \textbf{AliTok}~\citep{wu2025alitok}.  
For \emph{text-to-image} generation, we use \textbf{LuminaGPT2}~\cite{xin2025luminamgpt}, a recently released large-scale AR model that achieves the best performance on both GenEval~\cite{ghosh2023geneval} and DPG-Bench~\citep{hu2024ella}.  
All models adopt a VQ-style discrete tokenizer and support classifier-free guidance (CFG) at inference. 
Our method, \emph{SoftCFG}, is applied only at inference, without retraining or modifying the architecture.

\paragraph{Metrics.}
For \emph{class-conditional} generation task, we generate $50\mathrm{k}$ samples on the validation label set and report Fréchet Inception Distance (FID; lower is better). 
For completeness, we also track Inception Score (IS), precision/recall for generative models.

\paragraph{Hyperparameter settings.}
We follow the baseline sampling setup and keep all decoding hyperparameters identical to CFG (temperature, guidance scale $\gamma$, \etc). 
SoftCFG uses token-wise uncertainty weights $w_i = 1 - p_{\max}(x_i)$ measured on the \emph{unconditional} branch at the time token $x_i$ is generated, and perturbs only the unconditional value cache as defined in Sec.~\ref{sec:softcfg}. 
Guidance scale $\gamma$ matches the CFG baseline unless stated otherwise.

\subsection{Implications and Time Complexity Analysis}

We summarize the inference procedure of SoftCFG with step normalization in Alg.~\ref{alg:softcfg}. 
At each decoding step, the conditional logits are computed as in vanilla CFG, while the unconditional branch is temporarily perturbed by rescaling its value-cache entries according to token-wise confidence. 
Step normalization ensures that the perturbation weights are re-normalized at every step, allocating a fixed perturbation budget across the context. 
\textbf{Importantly, the perturbation is applied by an in-place scalar rescaling of the stored unconditional $\mathbf{V}$-cache, which is negligible compared to attention/MLP compute;} As illustrated in Fig.~\ref{fig:time_complexity},  the SoftCFG sampling time matches CFG.

\subsection{State-of-the-art Comparisons.}
\paragraph{Conditional Image Generation.}
\begin{table*}[t]
\vspace{-0.5cm}
\centering
\caption{ImageNet-1K~\citep{deng2009imagenet} 256 × 256 generation results evaluated with ADM~\citep{dhariwal2021diffbeatgan}. * indicates results from our own implementation.}
\label{tab:imagenet_results}
\resizebox{\textwidth}{!}{
\begin{tabular}{l|llrccccc}
\toprule
\toprule \textbf{Type} & \textbf{Generator} & \textbf{Venue} &\textbf{\#Params} & \textbf{FID}↓  & \textbf{IS}↑ & \textbf{sFID}↓& \textbf{Pre.↑} & \textbf{Rec.↑} \\
\midrule

 \multirow{11}{*}{Diff.} & LDM-8~\cite{rombach2022ldm} &CVPR~22 & 258M & 7.76 & 209.5 & - & 0.84 & 0.35 \\

 & LDM-4~\cite{rombach2022ldm} &CVPR~22 & 400M & 3.60 & 247.7 & - & \textbf{0.87} & 0.48 \\

 & UViT-L/2~\cite{bao2023all} & CVPR~23 & 287M & 3.40 & 219.9 & - & 0.83 & 0.52 \\
 & UViT-H/2~\cite{bao2023all} & CVPR~23 & 501M & 2.29 & 263.9 & - & 0.82 & 0.57 \\
 & DiT-L/2~\cite{peebles2023dit} & ICCV~23 & 458M & 5.02 & 167.2 & - & 0.75 & 0.57 \\
 & DiT-XL/2~\cite{peebles2023dit} & ICCV~23 & 675M & 2.27 & 278.2 & 4.60 & 0.83 & 0.57 \\
 & SiT-XL~\cite{ma2024sit} & ECCV~24 & 675M & 2.06 & 270.3 & 4.50 & 0.82 & 0.59 \\
 & DiMR-XL/2R~\cite{liu2024alleviating} & NeuIPS~24 & 505M & 1.70 & 289.0 & - & 0.79 & 0.63 \\
 & MDTV2-XL/2~\cite{gao2023masked} & ICCV~23  & 676M & 1.58 & 314.7 & 4.52 & 0.79 & 0.65 \\
 & REPA~\cite{yu2024repa} & ICLR~25  & 675M & 1.42 & 305.7 & 4.70 & 0.80 & 0.65 \\
 & REPA-E~\cite{leng2025repa-e} & ICCV~25  & 675M & \textbf{1.26} & \textbf{314.9} & 4.11 & 0.79 & \textbf{0.66} \\

\hline

 \multirow{4}{*}{Mask.} & MaskGIT~\cite{chang2022maskgit} & CVPR~22 & 177M & 6.18 & 182.1 & - & 0.80 & 0.51 \\
 & TiTok-S-128~\cite{yu2024titok} & NeuIPS~24 & 287M & 1.97 & 281.8 & - & - & - \\
 & MAGVIT-v2~\cite{yu2023magvit} &CVPR~23 & 307M & 1.78 & 319.4 & - & - & -\\
 & MaskBit~\cite{weber2024maskbit} &Arxiv~24 & 305M & \textbf{1.52} & \textbf{328.6} & - & - & - \\
\hline

\multirow{2}{*}{VAR} & VAR-d30~\cite{tian2024var} &NeuIPS~24 & 2.0B & 1.92 & 323.1 & - & 0.82 & 0.59 \\
& VAR-d30-re~\cite{tian2024var} &NeuIPS~24 & 2.0B & \textbf{1.73} & 325.0 & - & 0.82 & 0.60 \\
\hline

\multirow{3}{*}{MAR} & MAR-B~\cite{li2024mar} &NeuIPS~24 & 208M & 2.31 & 281.7 & - & 0.82 & 0.57 \\
& MAR-L~\cite{li2024mar} &NeuIPS~24& 479M & 1.78 & 296.0 & - & 0.81 & 0.61 \\
& MAR-H~\cite{li2024mar} &NeuIPS~24 & 943M & 1.55 & 303.7 & - & 0.81 & 0.62 \\
\hline

\multirow{2}{*}{FlowAR} & FlowAR-S~\cite{ren2024flowar} &ICML~25 & 170M & 3.61 & 234.1 & - & 0.83 & 0.50 \\
& FlowAR-H~\cite{ren2024flowar} &ICML~25 & 1.9B & 1.65 & 296.5 & - & 0.83 & 0.60 \\
\hline

\multirow{15}{*}{AR} & GPT2~\cite{esser2021taming} & CVPR~21 & 1.4B & 15.78 & 74.3 & - & - & - \\
& GPT2-re~\cite{esser2021taming} & CVPR~21 & 1.4B & 5.20 & 280.3 & - & - & - \\

& VIM-L~\cite{yu2021vim} & ICLR~22 & 1.7B & 4.17 & 175.1 & - & - & -\\
& VIM-L-re~\cite{yu2021vim} & ICLR~22 & 1.7B & 3.04 & 227.4 & - & - & -\\

& Open-MAGVIT2-B~\cite{luo2024openmagvit} & Arxiv~24 & 343M & 3.08 & 258.3 & - & \textbf{0.85} & 0.51 \\
 & Open-MAGVIT2-XL~\cite{luo2024openmagvit} & Arxiv~24 & 1.5B & 2.03 & 286.0 & - & 0.84 & 0.54 \\

& LlamaGen-L~\cite{sun2024llamgen} & Arxiv~24 & 343M & 3.07 & 256.1 & - & 0.83 & 0.52 \\
& LlamaGen-3B~\cite{sun2024llamgen} & Arxiv~24 & 3.1B & 2.18 & 263.3 & - & 0.81 & 0.58 \\
& RandAR-L~\cite{pang2024randar} & CVPR~25 & 343M & 2.55 & 288.8 & - & 0.81 & 0.58 \\
& RandAR-XXL~\cite{pang2024randar} & CVPR~25 & 1.4B &2.15 & 322.0 & - & 0.79 & 0.62 \\

& RAR-B~\citep{yu2024rar} & Arxiv~24 & 261M & 1.95 & 290.5 & - & 0.82 & 0.58 \\

&RAR-XL~\citep{yu2024rar} & Arxiv~24 & 955M & 1.50 & 306.9 & - & 0.80 & 0.62 \\

& Alitok-B~\citep{wu2025alitok} &Arxiv25 & 177M &1.50 &305.9 &- &0.78 &0.64  \\
& Alitok-L~\citep{wu2025alitok} &Arxiv25 & 318M &1.42 &\textbf{326.6} &- &0.78 &0.65 \\
& Alitok*-XL~\citep{wu2025alitok} &Arxiv25 & 662M &1.37	&321.4	&7.29	&0.79 &0.64  \\

\rowcolor{gray!20}& Alitok-B*+SoftCFG~(ours) &- & 177M &1.40	&271.0	&5.95	&0.78	&0.66 \\
\rowcolor{gray!20}& Alitok-L*+SoftCFG~(ours) &-  & 318M &1.39	&272.3	&\textbf{6.00}	&0.78	&\textbf{0.66} \\
\rowcolor{gray!20} & Alitok-XL*+SoftCFG~(ours) &- & 662M & \textbf{1.27}	&302.4	&6.76 &0.78	&0.65 \\
\bottomrule
\end{tabular}
}
\end{table*}
Table~\ref{tab:imagenet_results} summarizes results on ImageNet-1K $256{\times}256$.  Diffusion and masking models achieved strong IS and better FID.
Recent AR models (\eg, LlamaGen~\cite{sun2024llamgen}, RandAR~\cite{pang2024randar}) reduce this gap.
Our AliTok+SoftCFG attains an FID of 1.27, narrowing the gap between AR and diffusion models, while maintaining competitive IS and improved recall. With SoftCFG, AR demonstrates stronger potential to replace diffusion in future architectural designs.

\subsection{Ablations}
\label{sec:ablations}
\begin{table}[t]
\centering
\caption{Ablation study on ImageNet $256{\times}256$ class-conditional generation. We use the optimal $\gamma=13$~(in Eq.~\ref{eq:cfg}) and $k=1.4$~(in Sec.~\ref{sec:exp_hyperparameter}) for CFG on Alitok-XL~\cite{wu2025alitok}.
We report FID, IS and sFID. 
SoftCFG improves FID over vanilla CFG, and step normalization further stabilizes IS and sFID. Opt. indicates that we jointly tunes $(\gamma, k)$ via a small grid. We bold the top-2 values for each metric.}
\vspace{+0.3cm}
\label{tab:ablation}
\begin{tabular}{llll}
\toprule
\toprule
Method & FID $\downarrow$ & IS $\uparrow$ & sFID $\downarrow$\\
\midrule
Baseline &1.76 &221.2 &\textbf{5.55} \\
\phantom{0} + CFG &1.37~({\color{red}-0.39}) &\textbf{321.4}~({\color{red}+99.8}) &	7.29~({\color{green}+1.74}) \\
\phantom{0} + SoftCFG& \textbf{1.32}~({\color{red}-0.44})	&288.1~({\color{red}+66.9})	&7.62~({\color{green}+2.07}) \\
\phantom{0} + SoftCFG + StepNorm & \textbf{1.32}~({\color{red}-0.44})	&302.0~({\color{red}+80.8})	&7.16~({\color{green}+1.61}) \\
\phantom{0} + SoftCFG + StepNorm + Opt. & \textbf{1.27}~({\color{red}-0.49})	&\textbf{302.4}~({\color{red}+81.2})	&\textbf{6.70}~({\color{green}+1.15}) \\
\bottomrule
\end{tabular}
\end{table}
\paragraph{Effect of SoftCFG}
As shown in Table~\ref{tab:ablation}, replacing vanilla CFG with SoftCFG reduces FID (1.32 vs. 1.37), indicating more effective guidance.
We also show the visualization comparison on LuminamGPT2~\citep{xin2025luminamgpt} in~Fig.~\ref{fig:comparison_cfg_softcfg}.
However, as discussed in Sec.~\ref{sec:stepnorm}, directly applying SoftCFG could make the unconditional branch lose too much contextual information, therefore SoftCFG slightly destabilizes IS due to amplified step-wise perturbations.
\paragraph{Effect of Step Normalization}
Adding step normalization on top of SoftCFG mitigates the above issue, yielding both stable IS and improved sFID (7.16). This confirms that step normalization complements SoftCFG by regularizing its perturbation strength over time.
Figure~\ref{fig:step_norm_entropy} illustrates the normalized entropy on AliTok-XL with $\gamma=13$, $k=14$, 
where the confidence score is defined as the conditional probability after guidance.

\paragraph{Hyperparameter Varying}
\label{sec:exp_hyperparameter}
\begin{figure}
    \centering
\includegraphics[width=\linewidth]{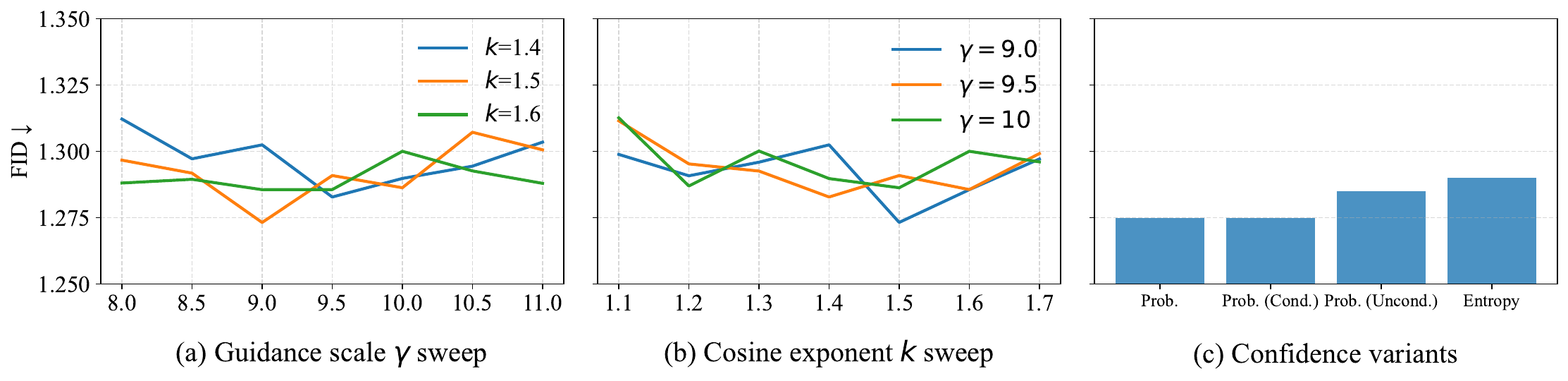}
\vspace{-0.5cm}
    \caption{Adaptive hyperparameter ablations of SoftCFG. 
    (a) Guidance scale $\gamma$ sweep across different cosine exponents $k$. 
    (b) Cosine exponent $k$ sweep under fixed guidance scales $\gamma$. 
    (c) Comparison of different confidence score definitions. 
    Results show that SoftCFG is robust to hyperparameter choices and that the max probability of conditional branch yields the best trade-off.%
    }
    \label{fig:paravaring}
\end{figure}
We analyze the sensitivity of SoftCFG to (i) the guidance scale $\gamma$~(in Eq.~\ref{eq:cfg} and Eq.~\ref{eq:softcfg}) and (ii) the power parameter $k$ in cosine scheduling introduced by \cite{yu2024rar}. 
Here, a cosine schedule for the guidance scale is: 
$\gamma_t = (\gamma-1)\cdot \tfrac{1}{2}(1-\cos((t/T)^k\pi))$, 
where $t$ is the current step and $k$ controls how guidance is distributed: larger $k$ shifts it to later steps, smaller $k$ to earlier steps. 
As shown in Fig.~\ref{fig:paravaring}~(a) and Fig.~\ref{fig:paravaring}~(b), SoftCFG consistently improves FID across a wide range of $\gamma$, while vanilla CFG quickly deteriorates for large scales. 
For cosine scheduling, moderate powers ($k=1.5$ or $1.6$) yield the best trade-off, whereas extreme values hurt performance.
These results highlight that \textbf{tuning is still required, and the effective ranges of $\gamma$ and $k$ are broader and the optimal values are typically smaller than those of vanilla CFG.}
This is because SoftCFG generally produces stronger guidance magnitudes than vanilla CFG.
In addition, Fig.~\ref{fig:paravaring}~(c) compares different definitions of the confidence score. 
We find that using the max probability of the conditional branch yields the most stable improvements, further showing that SoftCFG is robust to confidence variations.

\section{Conclusion}
In this work, we introduced SoftCFG, a lightweight modification to classifier-free guidance for autoregressive generation. Our key insight is that, similar to class tokens, previously generated visual content can also provide guidance to subsequent tokens. To fully exploit this effect, SoftCFG applies a soft weighting mechanism that adaptively modulates the influence of past tokens. While in this work we instantiated the weights with token-level confidence, our framework naturally accommodates more sophisticated scoring functions, including alternative scoring mechanisms, such as learned discriminators or perceptual evaluators. Extensive experiments on class-conditional generation demonstrate that SoftCFG consistently improves fidelity and alignment without retraining or increasing inference cost. We believe this simple yet general principle opens new directions for integrating fine-grained guidance signals into autoregressive generation.

\newpage
\section*{Ethics\&Reproducibility Statement}
This work relies only on public datasets, and while generative models may be misused, our method does not increase such risks.

All key settings are reported, and we will release core code for SoftCFG after the review process.

\bibliography{iclr2026_conference}
\bibliographystyle{iclr2026_conference}

\appendix

\newpage

\begin{figure}
    \centering
    \includegraphics[width=1\linewidth]{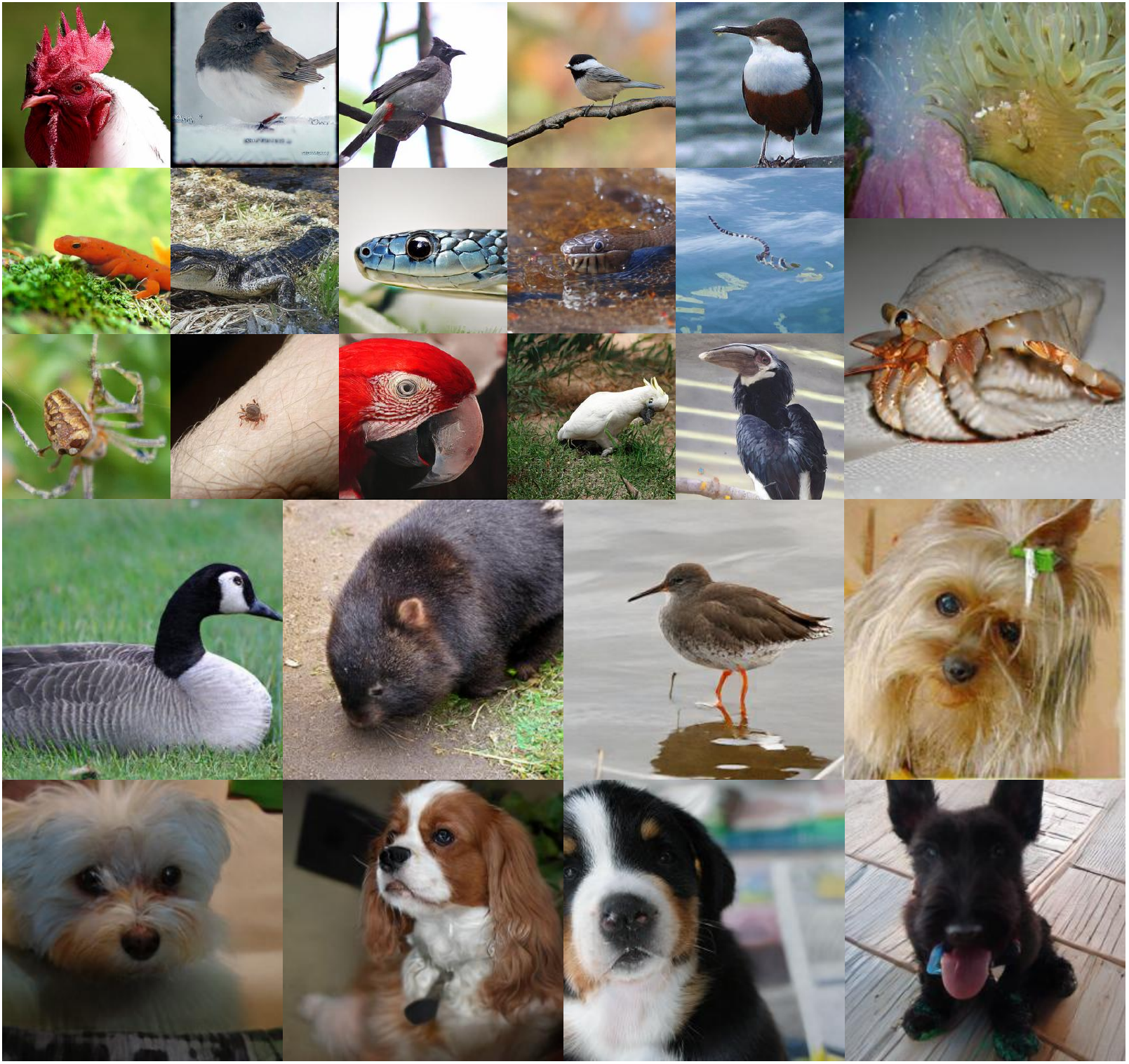}
    \caption{Samples generated by AliTok-XL with SoftCFG and Step Normalization on ImageNet $256\times256$. The results demonstrate high visual fidelity and semantic consistency across diverse classes.}
    \label{fig:alitok_demo}
\end{figure}
\section{Appendix: Visualization}
We also provide qualitative samples generated by AliTok-XL with SoftCFG and Step Normalization on ImageNet $256\times256$ (Fig.~\ref{fig:alitok_demo}). 
These results illustrate the visual diversity and semantic fidelity that can be achieved with our method, complementing the quantitative evaluations in Sec.~\ref{sec:ablations}.

\section{Appendix: Related Work}

\subsection{Visual AR Models}
Recent advances have significantly improved AR-based generation by enhancing both the visual tokenizer~\citep{yu2024titok, wu2025alitok,li2024imagefolder, qu2025tokenflow,shi2024IBQ} and the generation paradigm~\citep{tian2024var, pang2024randar, yu2024rar, xu2025daar, wang2025parallelized}, leading to sharper spatial alignment and more coherent semantics.
As a result, AR models~\citep{wu2025alitok, xu2025daar} have recently demonstrated generation quality on par with state-of-the-art diffusion-based and flow-based models~\citep{yu2024repa, leng2025repa-e} on standard benchmarks such as ImageNet~\citep{deng2009imagenet}.

\subsection{Classifier-free Guidance in Generative Models}
Classifier-Free Guidance (CFG) was first introduced in diffusion models to improve conditional generation performance without relying on external classifiers. It interpolates between conditional and unconditional predictions during sampling, trading off between sample fidelity and diversity~\citep{ho2022cfg}. Since then, CFG has become a fundamental technique in generative modeling.

A growing body of research explores ways to refine or extend CFG, particularly by injecting perturbations or structural modifications to guidance:
Self-Attention Guidance~\citep{hong2023sag} proposes using attention-based modifications to guide generation in diffusion models, improving sample quality without needing new objectives. This work highlights the importance of internal structure manipulation as an alternative to direct logits-based guidance.
Perturbed-Attention Guidance~\citep{ahn2024self} also introduces perturbed-attention into diffusion sampling to correct misaligned attention signals.
Smoothed Energy Guidance~\citep{hong2024seg} further addresses unstable energy landscapes in CFG by smoothing the attention curvature, resulting in more consistent guidance effects across different prompts and steps.
Guiding a Diffusion Model with a Bad Version of Itself~\citep{karras2024badguiding} shows that strong generation quality can be achieved by replacing the unconditional branch in CFG with a smaller or less-trained model. 

Several recent works ~\citep{shen2024saCFG, rajabi2025tpg, li2025acfg} also focus on improving CFG via dynamic and token-level perturbations:
Semantic-aware CFG~\citep{shen2024saCFG}, which adjusts guidance based on spatial semantic regions to mitigate inconsistency.
Token Perturbation Guidance~\citep{rajabi2025tpg}, which locally perturbs input tokens or hidden states to encourage more robust generation paths.
Adaptive CFG~(A-CFG)~\citep{li2025acfg}, which applies dynamic low-confidence masking to the unconditional branch based on step-wise uncertainty.

While these approaches offer significant improvements for diffusion and masked generative models, they are not directly transferable to AR generation, where conditioning signals degrade rapidly due to the sequential nature of decoding.
One of the few works to examine CFG in AR setting is Condition Contrastive Alignment~(CCA)~\citep{chen2024cca}, which enables guidance-free sampling by fine-tuning the model to align conditional and unconditional outputs.
However, it still underperforms standard CFG in most cases and requires additional training.
In contrast, our method introduces SoftCFG, a lightweight uncertainty-aware guidance mechanism for AR models, designed to stabilize the guidance signal across the entire generation sequence without modifying training, architecture, or inference efficiency. Beyond improving AR inference quality, SoftCFG provides a more stable and informative form of conditional guidance, which can potentially benefit future alignment-based or distillation-based approaches such as CCA and Model-guidance~\citep{tang2025diffwocfg}.

\section{Appendix: Normalized Entropy Computation}
\label{app:entropy}

Given the logits $\mathbf{z}_t \in \mathbb{R}^V$ at decoding step $t$, we first compute the probability distribution via softmax:
\begin{equation}
p_t(i) = \frac{\exp(z_{t,i})}{\sum_{j=1}^V \exp(z_{t,j})}, \quad i=1,\dots,V,
\end{equation}
where $V$ is the vocabulary size. 
The entropy of this distribution is
\begin{equation}
H(p_t) = - \sum_{i=1}^V p_t(i)\,\log p_t(i).
\end{equation}

To make entropy values comparable across vocabularies of different sizes, we normalize by the maximum entropy $\log V$:
\begin{equation}
\hat{H}(p_t) = \frac{H(p_t)}{\log V}, \qquad \hat{H}(p_t) \in [0,1].
\end{equation}

Here, $\hat{H}(p_t) \approx 0$ indicates highly confident predictions, while $\hat{H}(p_t) \approx 1$ corresponds to high uncertainty.

\section{Appendix: Derivation of the SoftCFG Deviation Bound}
\label{app:cfg_bound}

In this appendix, we derive the bound on the deviation of SoftCFG from vanilla CFG, given by:
\begin{equation}
\|\Delta^{\text{context}}_t\| \leq L_t \cdot \sum_{i=1}^{t-1} (1 - w_i) \|\mathbf{v}_i^{\text{uncond}}\|,
\end{equation}
where \(\Delta^{\text{context}}_t = \tilde{\mathbf{z}}^{\text{uncond, pertcontext}}_t - \mathbf{z}^{\text{uncond}}_t\) is the context-aware perturbation, and \(L_t\) is the Lipschitz constant of the model \(f_\theta\) with respect to the value cache.

\subsection{Perturbed Value Cache}
In SoftCFG, the unconditional value cache \(\mathbf{V}_{<t}^{\text{uncond}} = [\mathbf{v}_1^{\text{uncond}}, \mathbf{v}_2^{\text{uncond}}, \dots, \mathbf{v}_{t-1}^{\text{uncond}}]\) is perturbed as:
\begin{equation}
\tilde{\mathbf{V}}_{<t}^{\text{uncond}} = \mathbf{W}_{<t} \odot \mathbf{V}_{<t}^{\text{uncond}},
\end{equation}
where \(\mathbf{W}_{<t} = \text{diag}(w_1, w_2, \dots, w_{t-1})\), \(w_i = 1 - p_{\max}(x_i) \in [0, 1]\), and \(\odot\) denotes element-wise scaling. Thus:
\begin{equation}
\tilde{\mathbf{v}}_i^{\text{uncond}} = w_i \cdot \mathbf{v}_i^{\text{uncond}}, \quad i = 1, 2, \dots, t-1.
\end{equation}
The unconditional logits are computed as:
\begin{equation}
\mathbf{z}^{\text{uncond}}_t = f_\theta(\mathbf{K}_{<t}^{\text{uncond}}, \mathbf{V}_{<t}^{\text{uncond}}, x_{t-1}, \emptyset),
\end{equation}
\begin{equation}
\tilde{\mathbf{z}}^{\text{uncond, pertcontext}}_t = f_\theta(\mathbf{K}_{<t}^{\text{uncond}}, \tilde{\mathbf{V}}_{<t}^{\text{uncond}}, x_{t-1}, \emptyset).
\end{equation}
The deviation \(\Delta^{\text{context}}_t\) arises from the perturbation \(\tilde{\mathbf{V}}_{<t}^{\text{uncond}} - \mathbf{V}_{<t}^{\text{uncond}}\).

\subsection{Lipschitz Continuity}
We assume that \(f_\theta\) is \(L_t\)-Lipschitz continuous with respect to the value cache:
\begin{equation}
\|\tilde{\mathbf{z}}^{\text{uncond, pertcontext}}_t - \mathbf{z}^{\text{uncond}}_t\| \leq L_t \cdot \|\tilde{\mathbf{V}}_{<t}^{\text{uncond}} - \mathbf{V}_{<t}^{\text{uncond}}\|.
\label{eq:19}
\end{equation}

This assumption is justified by the structure of Transformer models, which consist of multi-head attention (MHA), feed-forward networks (FFN), residual connections, and LayerNorm:

\begin{itemize}
    \item \textbf{Attention Mechanism}: The attention output is computed as \(\text{Attention}(\mathbf{Q}, \mathbf{K}, \mathbf{V}) = \text{softmax}\left(\frac{\mathbf{Q} \mathbf{K}^T}{\sqrt{d_k}}\right) \mathbf{V}\). The softmax function is smooth with a Lipschitz constant of 1, as its Jacobian is bounded (\(\left|\frac{\partial \text{softmax}_i}{\partial x_j}\right| \leq 1\)). The value cache \(\mathbf{V}\) contributes linearly, and assuming bounded key and query matrices, the attention mechanism is Lipschitz continuous.
    \item \textbf{Feed-Forward Network}: The FFN, \(\text{FFN}(\mathbf{x}) = \sigma(\mathbf{W}_1 \mathbf{x} + \mathbf{b}_1) \mathbf{W}_2 + \mathbf{b}_2\), uses activation functions like ReLU or GELU, which are Lipschitz continuous.
    \item \textbf{Residual Connections and LayerNorm}: Residual connections (\(\mathbf{x} + \text{Attention}(\mathbf{x})\)) and LayerNorm have bounded gradients.
    \item \textbf{Composition}: The composition of Lipschitz continuous functions (attention, FFN, LayerNorm, and final linear projection) is itself Lipschitz continuous.
\end{itemize}

Form Eq.~\ref{eq:19}, we have:
\begin{equation}
\|\Delta^{\text{context}}_t\| \leq L_t \cdot \|\tilde{\mathbf{V}}_{<t}^{\text{uncond}} - \mathbf{V}_{<t}^{\text{uncond}}\|.
\label{eq:20}
\end{equation}

\subsection{Value Cache Difference}
The norm is:
\begin{equation}
\|\tilde{\mathbf{V}}_{<t}^{\text{uncond}} - \mathbf{V}_{<t}^{\text{uncond}}\| = \left\| \sum_{i=1}^{t-1} (1 - w_i) (-\mathbf{v}_i^{\text{uncond}}) \right\|.
\end{equation}
Applying the triangle inequality:
\begin{equation}
\left\| \sum_{i=1}^{t-1} (1 - w_i) (-\mathbf{v}_i^{\text{uncond}}) \right\| \leq \sum_{i=1}^{t-1} \left\| (1 - w_i) (-\mathbf{v}_i^{\text{uncond}}) \right\|.
\end{equation}
Since \(1 - w_i \geq 0\):
\begin{equation}
\left\| (1 - w_i) (-\mathbf{v}_i^{\text{uncond}}) \right\| = (1 - w_i) \|\mathbf{v}_i^{\text{uncond}}\|.
\end{equation}
Thus:
\begin{equation}
\|\tilde{\mathbf{V}}_{<t}^{\text{uncond}} - \mathbf{V}_{<t}^{\text{uncond}}\| \leq \sum_{i=1}^{t-1} (1 - w_i) \|\mathbf{v}_i^{\text{uncond}}\|.
\label{eq:24}
\end{equation}

\subsection{Final Bound}
Combining the Lipschitz condition in Eq.~\ref{eq:20} and the value cache difference in Eq.~\ref{eq:24}:
\begin{equation}
\|\Delta^{\text{context}}_t\| \leq L_t \cdot \sum_{i=1}^{t-1} (1 - w_i) \|\mathbf{v}_i^{\text{uncond}}\|.
\end{equation}
Thus, the deviation between SoftCFG and vanilla CFG is:
\begin{equation}
\|\mathbf{z}^{\text{SoftCFG}}_t - \mathbf{z}^{\text{CFG}}_t\| = \gamma \cdot \|\Delta^{\text{context}}_t\| \leq \gamma \cdot L_t \cdot \sum_{i=1}^{t-1} (1 - w_i) \|\mathbf{v}_i^{\text{uncond}}\|.
\end{equation}
This bound depends on the cumulative perturbation \(\sum_{i=1}^{t-1} (1 - w_i)\), which grows with sequence length \( t \), especially when many tokens have high confidence (\( w_i \to 0 \)). If \(\sum_{i=1}^{t-1} (1 - w_i)\) were bounded, the deviation would be \( O(\gamma) \), ensuring SoftCFG remains within a controlled trust region. However, without Step Normalization, this sum can become large, leading to excessive deviation and potential degeneration of the unconditional branch, as discussed in Sec.~\ref{sec:softcfg}.

\section{Appendix: Limitations and Future Work}
\begin{figure}[h]
    \centering
    \includegraphics[width=1\linewidth]{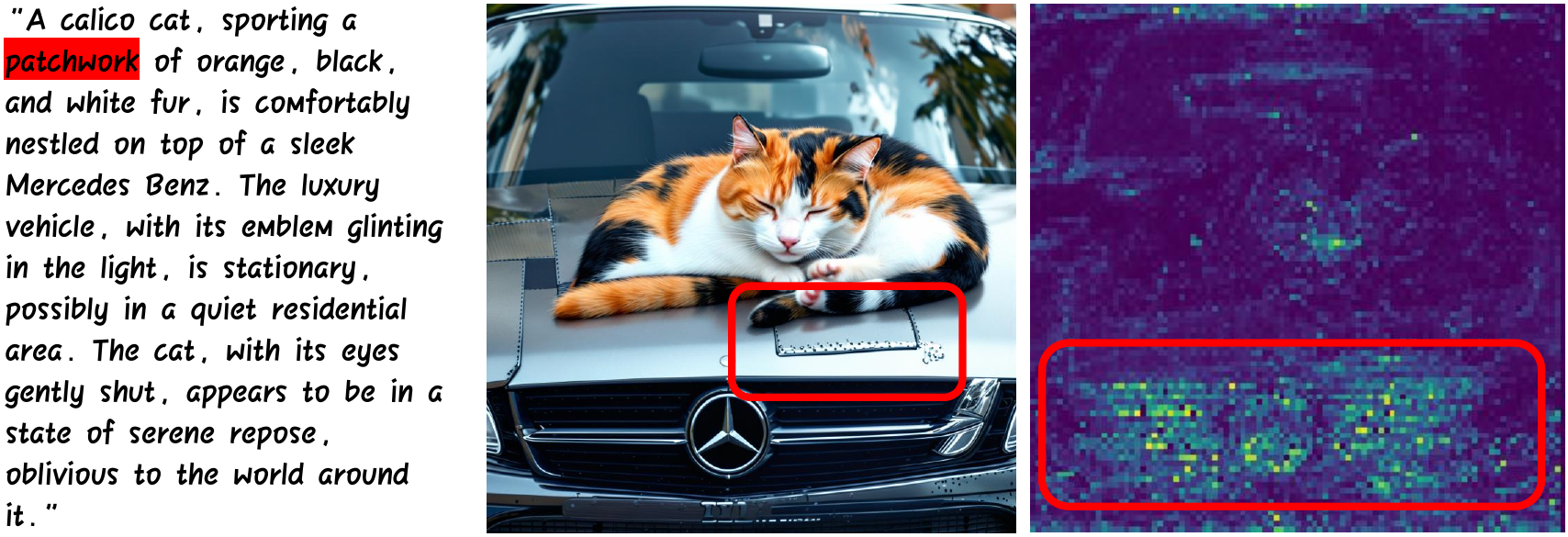}
\caption{
Failure case of SoftCFG. 
The prompt describes a calico cat on a Mercedes, but the ``patchwork'' attribute is wrongly applied to the car. 
Confidence focuses on the vehicle rather than the cat, leading to semantic drift.}

    \label{fig:failure_case}
\end{figure}

While SoftCFG improves the stability and effectiveness of classifier-free guidance in visual AR models, several limitations remain.

\textbf{First, the current instantiation relies on token confidence as the weighting function.} In complex cases, the confidence distribution may collapse onto a single object, leading to concept drift where generation is overly biased toward that object.
Figure~\ref{fig:failure_case} illustrates a representative failure mode of SoftCFG. 
Given the input prompt \textit{``A calico cat, sporting a patchwork of orange, black, and white fur, is comfortably nestled on top of a sleek Mercedes Benz...''}, the model generates an image where the ``patchwork'' attribute is incorrectly transferred to the car rather than the cat. 
The corresponding confidence map further shows that high-confidence regions are concentrated on the vehicle, while the cat receives low confidence. 
As a result, SoftCFG amplifies the vehicle features and neglects the cat, leading to semantic drift. 
This highlights that SoftCFG is sensitive to the accuracy of the confidence estimates, which current AR models cannot yet provide reliably.

\textbf{Second, is StepNorm too strict for SoftCFG?} While step normalization effectively prevents the deviation of SoftCFG from exploding, its strict renormalization may also be overly restrictive. 
By enforcing $\sum_{i=1}^{t-1}(1-\hat{w}_i)=1$ at every step, the perturbation budget is capped uniformly across sequence lengths. 
This implies that no matter how long the context grows, the total perturbation injected into the unconditional branch remains equivalent to at most a single token’s information. 
From another perspective, step normalization can be seen as discarding the cumulative contribution of multiple tokens and retaining only one token's worth of perturbation at each step. 
As a result, while it stabilizes long-horizon generation, its advantage in later steps may diminish compared to the unnormalized variant, see the orange line in Fig~\ref{fig:step_norm_entropy}, where a larger perturbation budget could potentially provide stronger guidance.

Future work could address these limitations: (1) a promising direction is to incorporate auxiliary perceptual models~(such as DINOV3~\cite{simeoni2025dinov3}) to score token-level semantics, yielding more robust confidence signals for guidance.  
(2) Although step normalization mitigates degeneration in long sequences, extremely long-horizon generation (\eg, high-resolution images or video) may still suffer from diminished guidance, calling for more adaptive normalization schemes.  
(3) Our study focuses primarily on class-conditional and text-to-image generation; broader applications such as multi-modal AR tasks, video synthesis, and reinforcement learning from human feedback (RLHF) remain unexplored.

\section{Appendix: LLM Usage}
During the preparation of this work, we made limited use of large language models (LLMs), specifically GPT, to assist with non-scientific tasks. These included (i) generating and formatting auxiliary visualization code (\eg, plotting scripts for ablation studies) and (ii) refining the writing style of the manuscript through grammar correction and phrasing suggestions. All scientific ideas, theoretical formulations, experimental design, and analysis were solely developed and validated by the authors.

\end{document}

%% file: math_commands.tex
\usepackage{amsmath,amsfonts,bm}

\def\eqref#1{equation~\ref{#1}}

\def\1{\bm{1}}

\DeclareMathAlphabet{\mathsfit}{\encodingdefault}{\sfdefault}{m}{sl}
\SetMathAlphabet{\mathsfit}{bold}{\encodingdefault}{\sfdefault}{bx}{n}

